# Artificial intelligence in drug discovery: A comprehensive review with a case study on hyperuricemia, gout arthritis, and hyperuricemic nephropathy

Junwei Su [a, †], Cheng Xin [a, †], Ao Shang [b], Shan Wu [c], Zhenzhen Xie [d], Ruogu Xiong [b], Xiaoyu Xu [b], Cheng Zhang [b], Guang Chen [b], Yau-Tuen Chan [b], Guoyi Tang [b, *], Ning Wang [b], Yong Xu [e], Yibin Feng [b, *]

[a] School of Computing and Data Science, Faculty of Engineering, The University of Hong Kong, Pokfulam, Hong Kong SAR, China
[b] School of Chinese Medicine, Li Ka Shing Faculty of Medicine, The University of Hong Kong, Pokfulam, Hong Kong SAR, China
[c] School of Resources and Environmental Engineering, Hefei University of Technology, Hefei, Anhui Province, China
[d] School of Management, Harbin Institute of Technology, Harbin, Heilongjiang Province, China
[e] School of Computer Science and Technology, Harbin Institute of Technology, Shenzhen, Guangdong Province, China
[†] Junwei Su and Cheng Xin contributed equally to this work
[*] Correspondence: tanggy11@hku.hk (Guoyi Tang); yfeng@hku.hku (Yibin Feng)

**Abstract**
This paper systematically reviews recent advances in artificial intelligence (AI), with a particular focus on machine learning (ML), across the entire drug discovery pipeline. Due to the inherent complexity, escalating costs, prolonged timelines, and high failure rates of traditional drug discovery methods, there is a critical need to comprehensively understand how AI/ML can be effectively integrated throughout the full process. Currently available literature reviews often narrowly focus on specific phases or methodologies, neglecting the dependence between key stages such as target identification, hit screening, and lead optimization. To bridge this gap, our review provides a detailed and holistic analysis of AI/ML applications across these core phases, highlighting significant methodological advances and their impacts at each stage. We further illustrate the practical impact of these techniques through an in-depth case study focused on hyperuricemia, gout arthritis, and hyperuricemic nephropathy, highlighting real-world successes in molecular target identification and therapeutic candidate discovery. Additionally, we discuss significant challenges facing AI/ML in drug discovery and outline promising future research directions. Ultimately, this review serves as an essential orientation for researchers aiming to leverage AI/ML to overcome existing bottlenecks and accelerate drug discovery.

## 1. Introduction

Drug discovery is a complex, multi-stage process aimed at identifying and providing drug candidates to be further assessed in clinical trials, through several critical stages including target identification, hit screening, lead optimization [1]. Given its importance in advancing healthcare, drug discovery is essential for addressing the growing need for treatments, particularly for diseases that currently have limited or no available therapies. However, modern drug discovery has faced significant challenges, such as unsustainable costs (often exceeding $2 billion per drug), high clinical attrition rates (often exceeding 90%), and lengthy timelines (often 10-15 years) [2-4]. This inefficiency stems from complex biological uncertainties, limitations of traditional screening methods, and the inherent difficulty of predicting efficacy and safety in humans [5]. AI, particularly ML, have emerged as powerful tools to address these challenges [6-8]. By leveraging tremendous biological, chemical, and clinical datasets, AI/ML are able to fundamentally revolutionize the drug discovery process, making it faster, more efficient, and more cost-effective [9-11].

In view of the progress made, there is an outstanding need for a comprehensive literature review of AI/ML applications in drug discovery. While there are many published review articles, they coincidentally have narrow scopes or just focus on specific stages, failing to provide an up-to-date holistic view of the entire pipeline [12-19]. As a result, they often overlook the dependence between various stages, missing important insights into how each phase influences the others [20-22]. Furthermore, some existing reviews concentrate on curating methods for each stage without offering concrete case studies, which are essential for understanding the practical applications and key challenges of these methods in the real-world drug discovery [23-25].

To address these gaps, this review comprehensively examines both conventional and emerging AI/ML techniques applied across the core stages of the drug discovery pipeline, including target identification (Section 2), hit screening (Section 3), and lead optimization (Section 4), as illustrated in **Figure 1**. This review aims to provide a concise overview and detailed insight into the key algorithms, models, solutions, advantages, and limitations of these techniques. Additionally, this review features a case study on drug discovery for hyperuricemia, gout arthritis, and hyperuricemic nephropathy, which significantly aggravate the global disease burden and urgently demand novel therapies to meet the unmet clinical needs [26-29]. The case study emphasizes the molecular targets and drug candidates identified using AI/ML approaches, highlighting the innovative opportunities in drug research and development for these diseases (Section 5). Inclusively, this review also discusses recent advances, promising success, key challenges, remaining obstacles, and forthcoming directions in this field (Section 6). This review may offer a comprehensive perspective on practical reality, transformative potential, and bright future of AI/ML in modern drug discovery.

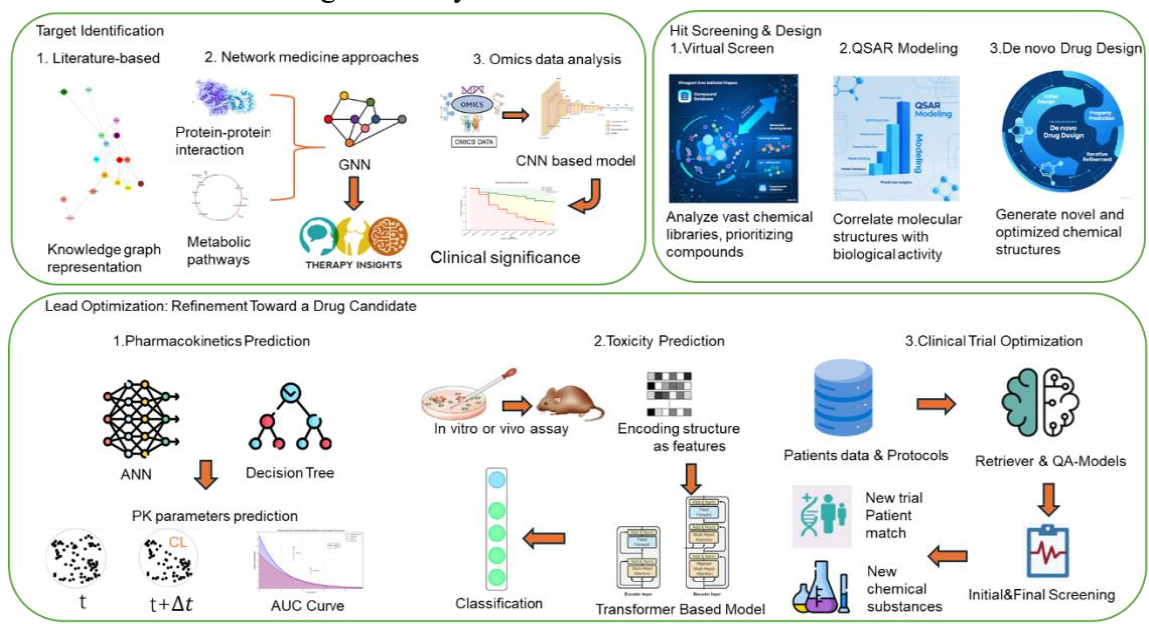

**Figure 1** AI/ML-driven innovative drug discovery.

## 2. AI/ML techniques in target identification and validation
### 2.1. Overview of target identification in drug discovery
Target identification and validation are fundamental early-stage processes in drug discovery, focusing on identifying and confirming biological entities—typically genes, proteins, or metabolites—that modulate disease mechanisms and thus serve as viable therapeutic targets. Traditionally, target identification was driven by academic research, which involved elucidating disease mechanisms, proposing candidate targets, and confirming their relevance through experimental validation [30-32]. This process often required extensive literature reviews and time-consuming experimental validation, frequently spanning so many years.

However, the rapid expansion of biomedical research has resulted in largescale genomic, proteomic, and other omics data, presenting both analytical challenges and vast opportunities [33]. These high-dimensional datasets are beyond human analytical capacity, making them ideally suited for ML methods, which excel at processing and interpreting complex, voluminous data [34]. As a result, ML has revolutionized target identification and validation, enabling more efficient and precise extraction of therapeutic targets from diverse biological datasets.

Formally, target identification can be defined as a computational task that aims to predict biologically relevant targets based on high-dimensional biological data and comprehensive biomedical knowledge bases. The effectiveness of ML in this domain heavily relies on the availability of diverse, high-quality datasets. The approaches used in target identification can be broadly categorized into the following sources:

- Biomedical Literature: Structured and unstructured text data used to mine disease–target associations.
- Network Biology: Interconnected biological datasets, such as PPIs, gene regulatory networks, and metabolic pathways, to uncover systemic relationships underlying disease mechanisms.
- Omics Data: Comprehensive biological measurements—including genomics, transcriptomics, proteomics, and metabolomics—to detect molecular signatures associated with diseases.

The following subsections provide an in-depth discussion of ML methodologies and applications within each of these categories.

## 2.2. Literature-based target identification

Literature-based target identification involves systematically mining biomedical texts—such as scientific publications, patents, and clinical reports—to discover potential therapeutic targets [35, 36]. Traditionally, this process relied on manual searches of databases like PubMed, followed by extracting and prioritizing disease-associated genes or proteins based on novelty, experimental evidence, and biological plausibility [37, 38]. However, these manual methods are limited by labor intensity, subjectivity, and a tendency to focus on well-established targets, making them inadequate for managing the ever-expanding biomedical literature [39].

Recent advances in ML, particularly NLP, have dramatically enhanced the efficiency and accuracy of literature-based target identification. ML-powered systems can quickly parse vast collections of biomedical texts, extracting relevant biological entities and their relationships [40]. Transformer-based NLP models like BERT, BioBERT, and SciBERT have improved the precision of biological interaction detection, unveiling previously overlooked disease–target associations and facilitating the identification of both novel and validated therapeutic candidates [41]. Additionally, AI-generated biomedical knowledge graphs integrate text-derived data with omics information, offering a multidimensional framework that supports hypothesis generation and systematic target validation [42].

Quantitative NLP-driven tools have further improved the scalability and objectivity of target evaluation. Bibliometric scoring systems such as TINX systematically quantify the novelty, relevance, and biological significance of potential therapeutic targets [43]. These systems leverage NLP techniques—such as entity recognition, relation extraction, and citation network analysis—to objectively rank candidate targets based on their interactions, research attention, and therapeutic potential. By incorporating bibliometric indicators like citation frequency and co-occurrence metrics, these tools significantly reduce bias, thereby accelerating decision-making and improving the reliability of target prioritization.

Several successful ML-driven applications in literature-based target identification underscore their practical impact:

- SciLinker: An NLP-based platform that identified over 1.25 million gene–disease associations, uncovering both established and novel therapeutic targets [44].
- BenevolentAI's Knowledge Graph: Integrated NLP with multiomics data to identify baricitinib as a potential COVID-19 therapy, validated through clinical trials [45].
- Rosalind: Used tensor factorization on literature-based knowledge graphs to predict and validate new therapeutic targets in rheumatoid arthritis research [46].
- Insilico Medicine's Clinical Progress: Combined literature mining with multi-omics data to identify a novel fibrosis target, resulting in the first AI-designed drug candidate entering clinical trials [47].

*2.3. Network medicine approaches for target identification*

Network medicine applies network science and systems biology principles to identify therapeutic targets by analyzing interconnected biological systems such as PPIs, gene regulatory networks, metabolic pathways, and disease–gene associations (**Figure 2**) [48-50]. Network medicine is based on the idea that diseases arise from disruptions in complex biological networks, rather than isolated genetic or molecular alterations. Therefore, identifying therapeutic targets requires analyzing how these disruptions propagate through interconnected molecular entities.

Traditional network-based methods identified potential therapeutic targets by analyzing their positions within biological networks [51, 52]. Common strategies such as guilt-by-association, network diffusion, centrality analysis, and module detection enabled the exploration and prioritization of targets based on their roles within these networks.

Historically, network-based target identification was limited by heuristic methods and manual analyses, leading to inefficiencies and scalability issues when confronted with complex, high-dimensional data [52, 53]. The integration of ML, particularly GNNs, has overcome these limitations, thus enhancing the scale and precision of network medicine.

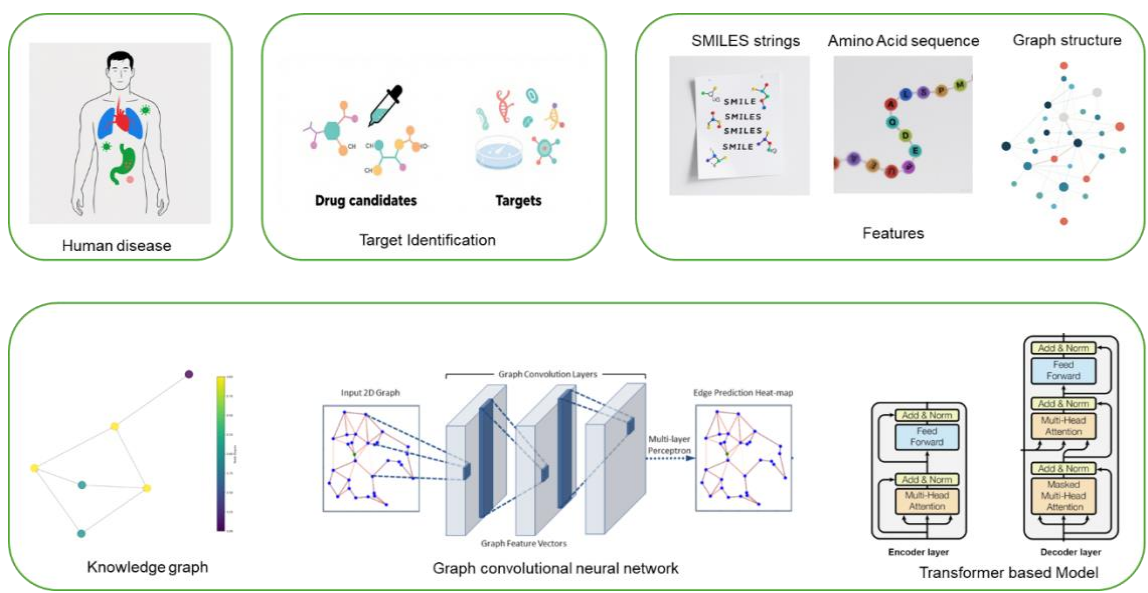

**Figure 2** AL/ML-aided novel network pharmacology.

Modern ML-driven network medicine techniques automate the analysis and integration of multilayered biological data, including genomics, proteomics, clinical phenotypes, and literature-derived evidence. GNN-based methods such as GCNs, GATs, and message-passing neural networks can identify complex, nonlinear relationships from network structures [54]. These approaches enable precise predictions of disease-specific network perturbations, facilitating the identification of high-confidence therapeutic targets [55, 56].

Recent ML-driven applications demonstrate the transformative impact of network medicine for target identification:
- deepDTnet: An integrated ML framework that uses heterogeneous biomedical networks to predict novel drug–target interactions, such as identifying RORt as a target for topotecan [57].
- Biomedical Knowledge Graph Embeddings: This approach integrates structured biological databases with literature-derived data through embedding methods, significantly improving predictive accuracy and revealing new therapeutic insights [58].
- Personalized Network Medicine: Using ML-driven analyses of individual molecular and clinical data, personalized disease networks were constructed to identify therapeutic targets tailored to specific patient profiles [57].

*2.4. Omics data analysis for target identification*
Omics data analysis leverages large-scale datasets from genomics, transcriptomics,

proteomics, and metabolomics to identify and prioritize potential therapeutic targets. Traditional methods for target identification typically involved manually analyzing isolated omics datasets to detect molecular abnormalities, which often resulted in fragmented insights and inefficiencies [59].

Recent advances in ML have revolutionized omics-based target identification. ML algorithms now process high-dimensional data more rapidly and accurately, detecting subtle, multi-layered biological patterns those traditional methods may overlook. Moreover, ML facilitates the integration of multiple omics datasets, combining genomic, transcriptomic, proteomic, and metabolomic information to provide a more comprehensive understanding of disease mechanisms [60].

Sophisticated ML techniques, including DNNs, CNNs, and autoencoders, have become essential in analyzing complex nonlinear relationships in omics data. By integrating omics data with biological networks and pathway analyses, these ML methods enhance the interpretability and predictive power of target discovery.

Notable recent ML-driven omics applications for target identification include:
- TNIK in Idiopathic Pulmonary Fibrosis: AI-based multi-omics platform identified TNIK as a novel therapeutic target, facilitating the development of a first-in-class inhibitor [47].
- FRoGS: A DL model using functional gene embeddings improved the accuracy of target prediction from transcriptomic data, highlighting novel candidate genes [61].
- Network-Based Multi-Omics Integration (AOPEDF and REMAP): AI-powered frameworks integrated multi-omics data with biological networks, uncovering previously unrecognized targets for therapeutic intervention [55, 56].
- AlphaFold-driven Omics Integration: Combining AlphaFold's protein structure predictions with genomics and proteomics data enabled rapid identification of structurally validated targets across diseases [62].

This shift toward ML-driven analysis is transforming omics-based target identification, enabling faster, more reliable identification of therapeutic targets for drug discovery.

**3. AI/ML techniques in hit screening and design**
*3.1. Overview of hit screening in drug discovery*
Hit identification and generation are critical early-stage processes in drug discovery that focus on discovering molecular compounds ("hits") with desirable biological activities against specific therapeutic targets [63]. Hit identification refers to computational and experimental methods used to systematically screen large chemical libraries and identify molecules likely to interact effectively with biological targets. In contrast, hit generation involves designing entirely novel molecules tailored for specific biological interactions, expanding beyond known chemical entities.

The success of hit identification and generation is crucial for the later stages of drug discovery, including lead optimization and preclinical development. Effective early-stage identification of promising molecular candidates increases the likelihood of progressing potent, selective, and safe therapeutic agents toward clinical application [64]. However, significant challenges exist in both processes. The vast chemical diversity—estimated to exceed 1060 possible small molecules—makes exhaustive exploration impractical and necessitates efficient computational strategies for navigating this immense chemical space [65]. Traditional experimental screening methods often struggle with scalability, especially when faced with increasingly extensive molecular libraries [66]. Moreover, reliably predicting bioactivity computationally remains challenging, requiring predictions that align closely with experimental validation. Additionally, ensuring the synthetic feasibility of computationally proposed molecules remains an ongoing hurdle, as many in silico hits may be challenging or resource-intensive to synthesize experimentally [67].

Recent advancements in ML have made significant strides in overcoming these challenges, improving the efficiency and accuracy of both hit identification and generation. ML-driven virtual screening methods efficiently analyze vast chemical libraries, prioritizing compounds based on predicted bioactivities and binding affinities. Similarly, ML-powered QSAR models correlate molecular structures with biological activity, improving prioritization and optimization of potential hits. Furthermore, advanced ML-based generative techniques have revolutionized de novo drug design, employing methods such as GANs, VAEs, diffusion models, and RL to generate novel, optimized chemical structures. These methods enable robust exploration of chemical diversity, accurate bioactivity prediction, and accelerated discovery of therapeutically viable compounds. Detailed descriptions of these methodologies are provided in the following subsections.

*3.2. Virtual screening*

Virtual screening is a key technique in hit identification and generation, used to computationally evaluate large chemical libraries and efficiently identify promising molecules ("hits") before experimental validation [68-70]. By narrowing down the initial pool of candidate compounds, virtual screening accelerates drug discovery and provides high-quality starting points for subsequent lead optimization. virtual screening methods are generally divided into two main categories: SBVS and LBVS, both of which have significantly benefited from recent advances in ML [71-73].

SBVS uses detailed three-dimensional information of biological targets—obtained from techniques like X-ray crystallography, Cryo-EM, or computational predictions such as AlphaFold—to evaluate potential ligand interactions [74-77]. Molecular docking, a key component of SBVS, predicts ligand binding poses and assesses interaction strengths. However, traditional docking scoring functions often suffer from limitations in accuracy, which can reduce their effectiveness in identifying active hits [78-80].

Recent advancements in ML have significantly enhanced SBVS by addressing these limitations:

- Enhanced Scoring Functions: ML-based scoring functions, including CNNs and graph-based models, significantly improve predictions of binding affinity and docking pose accuracy. These models leverage extensive structural datasets to reliably distinguish active compounds from inactive ones, improving hit identification quality [81-83].
- Ultra-Large Library Screening: ML-driven methods like Deep Docking use iterative, active-learning approaches to efficiently screen vast chemical libraries, prioritizing subsets of compounds based on predictive models and enhancing hit discovery [84-86].
- Integration with Hit Generation (Structure-Based Design): AI-driven methods, including GNNs and diffusion-based models, integrate virtual screening with de novo molecule generation. These techniques not only predict ligand conformations but also generate novel molecules designed for specific binding pockets, effectively linking hit identification and generation [87-89].

LBVS is another approach that complements structure-based methods, particularly when structural information of the biological target is limited. LBVS identifies active compounds based on known ligands and their properties, operating under the assumption that structurally similar molecules exhibit comparable biological activities. Traditional methods in LBVS include similarity searching, pharmacophore modeling, and QSAR analyses [90, 91].

ML has greatly enhanced LBVS by improving predictive capabilities and enabling efficient exploration of large chemical spaces:

- Advanced Similarity Search and Pharmacophore Modeling: ML algorithms, such as DNNs and embedding methods, improve the identification of structurally and functionally similar ligands. These models can capture complex structural relationships, improving the accuracy of similarity-based searches [92-94].
- Enhanced QSAR Models: Modern QSAR models use ML methods, including gradient boosting, RF, and DL architectures, to predict biological activities from molecular descriptors. These ML-enhanced QSAR models improve the hit prioritization and optimization in ligand-based screening processes [95-97].

In summary, virtual screening, both structure-based and ligand-based, has been significantly enhanced by AI-driven techniques, improving efficiency, accuracy, and integration. These advances allow for more comprehensive exploration of chemical space and the identification or generation of high-quality hits for further drug discovery efforts.

*3.3. QSAR modeling*

QSAR modeling plays a pivotal role in both hit identification and generation by linking chemical structures of molecules to their biological activities through computational descriptors and ML techniques. Traditionally, QSAR methods relied on linear regression and simple ML algorithms with manually curated descriptors, limiting their ability to capture complex relationships between molecular structure and biological activity [96, 98]. These methods often fell short in reliably predicting bioactivity and prioritizing hits for further validation.

Recent advances in ML have substantially improved QSAR modeling, enabling more accurate and comprehensive bioactivity predictions [99, 100]. Modern ML-driven QSAR approaches address key challenges, enhancing both hit identification and generation:

- Representation Learning: DNNs automatically learn meaningful molecular representations from raw structural data, such as SMILES strings or molecular graphs, eliminating reliance on predefined descriptors. This capability enhances the accuracy of biological activity predictions and streamlines hit identification [101-104].
- Multi-task and Transfer Learning: Contemporary QSAR models use multi-task learning to predict biological activities across multiple targets, leveraging shared chemical features. Transfer learning approaches, which pre-train models on large chemical datasets and then fine-tune them for specific targets, further boost model performance, accelerating hit identification and generative design [105-107].
- Hybrid QSAR Approaches: Integrating QSAR models with additional data sources, such as docking scores or physics-based descriptors, creates hybrid models that combine ligand- and structure-based methods. These hybrid approaches enhance predictive accuracy and reliability, supporting the identification of promising hits and guiding molecular optimization during hit generation [108-110].
- Model Interpretability and Trustworthiness: AI-driven QSAR models increasingly incorporate interpretability techniques, revealing key molecular features that influence biological activity. These advancements build chemists' confidence and inform rational optimization strategies during the iterative design of new molecules in hit generation [108-110].

In summary, modern ML-enhanced QSAR models have transformed drug discovery, significantly improving hit identification and generation by providing more accurate, interpretable, and actionable predictions for therapeutic development.

*3.4. De novo drug design*

De novo drug design focuses on creating entirely novel chemical structures with desired biological activities, expanding beyond the selection of existing molecules [111, 112]. While traditional methods like fragment-based growth and evolutionary algorithms were valuable, they often faced limitations related to chemical feasibility, synthetic accessibility, and structural complexity [113, 114].

Recent ML advancements have enhanced de novo drug design by enabling the rapid, systematic generation of chemically realistic and optimized molecules. Key ML-driven generative approaches include:

- VAEs: VAEs encode molecular structures into a continuous latent representation, enabling efficient exploration of chemical spaces and generation of molecules optimized for desired properties [115-117].
- GANs: GANs generate novel, chemically plausible molecules through adversarial training between a generator and a discriminator, refining molecules to ensure structural diversity and synthetic feasibility [118-120].
- Diffusion Models: These models progressively transform random noise into 3D molecular structures tailored to specific protein targets, directly integrating structure-based hit identification with molecular design [121-123].
- RL: Combining generative models with RL, this method iteratively optimizes molecular designs based on reward signals, ensuring newly generated molecules possess desirable properties and enhancing both hit identification and generation [124-126].

These ML-driven methods have moved beyond academic research into real-world applications, with AI-designed compounds progressing into clinical trials. Nevertheless, generated molecules still require filtering, synthetic feasibility assessment, and experimental validation to confirm their efficacy and practical utility. The integration of AI with expert chemical knowledge represents a significant step forward in expanding chemical spaces and facilitating the design of innovative therapeutic candidates.

**4. AI/ML techniques in lead optimization**

*4.1. Overview of lead optimization in drug discovery*

Lead optimization is a critical phase in the drug discovery pipeline which follows hit identification and lead generation. In this stage, initial chemical leads-molecules with promising biological activity are systematically modified to enhance their pharmacological profiles. The goal is to improve drug-likeness, including PK, toxicity prediction and clinical trial optimization. This step is essential for transforming a bioactive compound into a viable preclinical candidate. In the whole pipeline, lead optimization serves as the bridge between high-throughput screening and clinical development, ensuring that compounds meet essential criteria for in vivo and vitro efficacy, favorable ADME properties. Historically, this process relied on labor-intensive in vitro assays and empirical medicinal chemistry reactions. But now with the development of AI and ML techniques, the time and cost will be reduced via rapid and data-driven predictions of molecular behavior to improve the likelihood of success in downstream tasks.

Modern computational approaches range from classical algorithms such as RF, SVM and Gradient Boosting. To more advanced techniques such as GNN and Transformer-based architectures which will predict PK parameters, drug concentrations, ADME profiles, and toxicity endpoints, often outperforming traditional QSAR models. They are all beneficial to accelerating decision-making, reducing experiment cost and improving the success rate of candidate selection. We will introduce the application of these models in lead optimization in the following sections.

*4.2. PK prediction*

In the field of lead optimization, it is essential to identify the compound with fine drug-likeness, which could be used to understand and enhance the PK prediction, including the ADME properties. But traditional methods such as rule-based method and linear QSAR models in vitro and vivo are time consuming and expensive [127, 128]. Other issues, such as plasma concentration-time profiles and tailored dosing techniques, will be addressed. Thus, the prevalent AI and ML methods could be an efficient alternative, which will enable the early and accurate prediction in the ADME behavior to present more information to guide the medicinal chemists in the whole pipeline. The recent studies try to tackle the problem in the four following fields: (1) biologics and protein-based drug modeling, (2) traditional small molecule PK parameter prediction, (3) personalized PK modeling using patient data, (4) ADME and toxicology models, and (5) hybrid modeling and model fusion approaches.

ML has significantly advanced the prediction of PK parameters for biologics, such as insulin analogs and monoclonal antibodies, which are typically harder to model than small molecules due to their size and structural complexity. Einarson et al. used RF and ANN to accurately predict clearance and half-life for insulin analogs, leveraging features like molecular size and charge [129]. Similarly, Obrezanova et al. applied DL methods, including DNN and Gaussian Process Regression, to translate chemical features into PK parameters across species [130]. Krutilka Patidar et al. proposed a hybrid framework combining minimal PBPK modeling with ML classifiers to screen over 11,000 antibody candidates, reducing reliance on wet-lab experimentation [131].

A major application of ML in PK is the prediction of fundamental parameters—such as clearance (CL), volume of distribution (Vd), and AUC—for small molecule drugs. These models often use molecular descriptors and preclinical data to forecast human PK behavior. Kwapien et al. compared PLS, RF, and XGBoost for drug clearance modeling, concluding that ML better captures non-additive effects [132]. Kamiya et al. used LightGBM to estimate hepatic clearance and absorption rates for a diverse chemical set [133]. Iwata et al. introduced Deep Tensor and XGBoost for cross-species predictions of CL and Vd, integrating chemical and in vitro data [134, 135]. Classical methods were also extended by hybrid ML approaches, such as Destere et al. who improved isavuconazole CL prediction via XGBoost + Bayesian estimation [136]. Studies by Tang et al. and Keutzer et al. further validated that ML models can match or outperform traditional pharmacometric models, particularly in handling variability from neonatal and adult PK profiles [137, 138].

ML is transforming individualized dosing and PK modeling by incorporating patient-specific covariates. These models improve predictive accuracy for therapeutic drug monitoring and dosage optimization. Khusial et al. developed an ANN combining nonlinear mixed-effects modeling (NONMEM) model for olanzapine plasma concentrations, effectively using patient data such as age and hematocrit [139]. Ponthier et al. applied ML to predict tacrolimus AUC in transplant patients, emphasizing the importance of formulation-specific data like MeltDose [140]. Similarly, Hughes et al. used a hybrid XGBoost-Bayesian model to refine vancomycin dosing, adjusting Bayesian priors based on patient-level factors like serum creatinine [141]. Labriffe et al. demonstrated that XGBoost can handle sparse datasets for everolimus, a scenario where traditional methods often struggle [142].

In early drug development, ML supports in silico ADME (Absorption, Distribution, Metabolism, Excretion) and toxicity prediction, aiding in the rapid screening and optimization of drug candidates. Li et al. showed that LightGBM could predict anti-breast cancer ADME profiles with ~90% accuracy using features like solubility and lipophilicity [143]. An et al. demonstrated that tree-based models (XGBoost, LightGBM) outperform RF in early-stage screening [144]. Brereton et al. introduced POEM, a parameter-free method that improved ADME generalizability [145]. Wang et al. explored traditional Chinese medicine compounds using RF and SVM to identify promising candidates for rheumatoid arthritis [146]. Karalis combined PCA with RF to improve predictions of bioequivalence metrics (Cmax/Tmax), and Yang et al. applied transformer-based DL to optimize ADMET in drug design [147, 148]. Cobre et al. used a mix of PCA, ANN, and SVM to screen natural SARS-CoV-2 inhibitors, integrating chemical features for safety and efficacy [149].

Some of the most robust PK models arise from combining ML with classical pharmacometric frameworks, creating hybrid systems that benefit from both mechanistic interpretability and predictive power. Hughes et al. and Destere et al. incorporated Bayesian modeling with ML (e.g., XGBoost) to personalize vancomycin and isavuconazole dosing [141, 150]. Khusial et al. successfully blended ANN with NONMEM to enhance olanzapine predictions [139]. Krutilka Patidar et al. integrated minimal PBPK models with classification trees to enable high-throughput screening for antibodies [131]. Yang et al. introduced transformers to support sequence-sensitive learning in ADMET predictions, further showing that model fusion can significantly expand the scope and reliability of ML in PK [148].

To sum up, from preclinical drug development such as CL and Vd prediction to clinical precision dosing such as vancomycin, tacrolimus and AUC. XGBoost, ANN, and hybrid models are the most prevalent, but challenges remain in data scarcity, model interpretability, and real-world validation. Future directions include explainable AI for regulatory acceptance and federated learning for multi-institutional PK datasets.

*4.3. Toxicity prediction*

Toxicity prediction is critical in modern drug discovery, which serves as a gatekeeper to identify potentially harmful compounds before they reach clinical trials. Historically, toxicity assessments relied on in vivo animal models and in vitro assays, but they are time-consuming and inaccurate in predicting human toxicity due to interspecies [151]. Thus, the ML and DL methods could be an alternative due to its efficiency in predicting various types of drug-induced toxicity. Traditional algorithms such as RF, SVM, and gradient boosting have shown renowned performance in classifying toxic and non-toxic compounds based on molecular descriptors and fingerprints [152]. In particular, we can classify all the models into the (1) DL model for toxicity prediction and (2) explainable and generation modeling approaches for toxicity.

DL models for toxicity prediction encompass the core predictive frameworks developed for ADMET and toxicity modeling, using ML, DL, and graph-based methods. These models focus on accuracy, generalizability, and leveraging complex molecular inputs like SMILES strings or graph structures.

- MolToxPred: A stacked ensemble of RF, MLP, and LightGBM that achieved AUROC >85% across multiple datasets, showing robustness through ensemble learning [153].
- DeepTox: A multi-task DL model trained on chemical descriptors and SMILES strings that won the Tox21 Challenge by predicting toxicity across 12 assays [154].
- SMILES-Mamba: A transformer-based model using a pretraining–finetuning pipeline for high-performance ADMET prediction, with strong results on toxicity endpoints [155].
- Deep-PK: A unified platform combining ADMET and PK prediction, integrating analysis and optimization features to support early drug development. The platform applied GNN and molecular graph-level features to predict ADMET and toxicity, enabling structural reasoning over molecules [156].

Explainable and Generation modeling approaches for toxicity have demonstrated in several studies that advances interpretability (through explainable AI, or explainable AI) and non-traditional modeling paradigms (e.g., quantum-based neural networks) to address critical issues like transparency and cross-platform model transfer.
- Explainable AI: A review emphasizing the importance of explainable AI in drug safety modeling, particularly for regulatory acceptance of toxicity predictions [157].
- Quantum-based modelling: Proposed an innovative approach for transferring quantum-trained neural network weights to classical architectures for toxicity modeling, demonstrating accuracy retention and expanding the modeling toolkit [158].

Despite these advances, significant challenges persist. These include limited availability of high-quality, labeled toxicity datasets, the need for better generalizability across chemical space, and the integration of multi-modal data sources such as transcriptomics and histopathology. As highlighted by Masarone, S. et al., future research should emphasize hybrid modeling approaches that integrate in silico, in vitro, and in vivo data to enhance both the robustness and biological relevance of toxicity predictions [159].

In summary, recent studies reflects a rapid evolution in computational toxicology, with ML and DL techniques offering promising tools for early-stage toxicity screening. However, continued methodological innovation, improved data accessibility, and interpretability remain essential for the practical adoption of these models in real-world drug development pipelines.

*4.4. Clinical trial optimization*

Clinical Trial is the ultimate phase in the whole pipeline for the drug discovery, facing challenges such as slow patient recruitment, low accuracy in identifying novel candidates, suboptimal trial design and high failure rates. By introducing NLP techniques, we could classify the methods into the following fields: (1) NLP for biomedical knowledge extraction and trial design, (2) AI/ML for patient modeling and adaptive clinical trials, (3) regulatory automation and AI-augmented compliance and (4) PLMs for target discovery.

NLP is being leveraged to extract, organize, and harmonize biomedical knowledge from unstructured text sources such as scientific literature, patents, and clinical trial protocols. These applications significantly reduce redundancy, improve trail design, and enable strategic decision-making. By leveraging NLP, researchers can extract critical insights from scientific literature, patents, and experimental data to identify promising drug targets and therapeutic strategies [160]. These capabilities extend to intellectual property landscapes, where NLP techniques can assess prior art, identify novel chemical substances, and locate licensing or collaboration opportunities based on historical patient and research data [161]. In the context of clinical trials, NLP also plays a central role in harmonizing trial protocols by comparing and aligning elements from previous studies, helping researchers avoid design redundancy and maintain regulatory alignment [162]. Furthermore, the use of NLP to parse eligibility criteria from clinical trial registries and patient records has significantly accelerated patient cohort identification and eligibility screening, streamlining one of the most time-consuming stages of trial initiation [163]. Collectively, these developments position NLP as a powerful enabler for evidence-driven and efficient trial design.

AI, particularly ML and DL, have become instrumental in reshaping clinical trials by enabling patient-specific modeling and adaptive study designs. Through the analysis of EHRs, ML algorithms can uncover hidden patterns linking patient characteristics with disease progression and treatment outcomes, ultimately enabling the discovery of novel biomarkers and improving predictions of therapeutic response [164]. In parallel, AI methods such as RL and Bayesian optimization are now used to simulate dosing regimens in silico, allowing researchers to explore a wide range of therapeutic strategies and eliminate ineffective or unsafe ones before real-world testing begins [165]. This capability is especially impactful in high-stakes fields like oncology and neurology, where dosing precision is critical to balancing efficacy and safety [166]. Further advancing personalization, digital twins—virtual models of patients created from real-world health data—are being employed to simulate responses to treatments, making it possible to preemptively identify adverse effects and optimize outcomes for targeted populations [167]. Complementing these innovations, generative AI techniques have shown promise in synthesizing realistic clinical datasets, enabling better training of diagnostic and predictive models [168], with technologies like Super-Resolution GAN demonstrating utility in enhancing medical imaging and video diagnostics [169]. Together, these advancements mark a paradigm shift toward data-driven, individualized clinical trials.

The growing complexity of regulatory requirements in clinical research has catalyzed the integration of AI, particularly Large Language Models (LLMs), to streamline compliance and protocol development. LLMs are increasingly employed to interpret FDA guidance documents, automate the drafting of Institutional Review Board protocols, and generate compliant study designs—significantly reducing the time and manual effort typically required for regulatory submissions [170]. In tandem, tools such as ChatGPT Advanced Data Analysis are emerging as accessible platforms that democratize ML and data processing for medical researchers, helping non-experts conduct complex analyses, visualize findings, and develop regulatory-ready documentation with minimal technical overhead [171]. These tools not only expedite the development process but also enhance reproducibility and transparency, aligning well with evolving standards for AI-assisted healthcare innovation.

At the molecular level, the rise of transformer-based PLMs has opened new frontiers in drug discovery by enabling sophisticated protein-level understanding. Drawing inspiration from NLP, these models treat amino acid sequences like language, learning to infer structural and functional features directly from sequence data. Trained on vast protein databases like UniProt, PLMs such as those introduced are now capable of predicting protein folding, functional annotations, and even the effects of mutations without requiring crystal structures [74, 172, 173]. Advanced models like Evolutionary Scale Modeling-2 and ProtGPT2 further extend this capability by incorporating evolutionary and generative principles, enabling de novo protein design and interaction modeling [174, 175]. OmegaFold, in particular, incorporates 3D geometric awareness, allowing for more accurate structure prediction and application in spatially dependent drug–protein interactions [176]. These innovations bridge computational biology and drug development, positioning PLMs as core tools for rational drug design, target validation, and therapeutic protein engineering.

In sum, the convergence of NLP and AI technologies is laying the groundwork for a more adaptive, ethical, and efficient clinical trial infrastructure, combing the techniques from the Bayesian network and RL, NLP and generative AI, give us more insights on the drug synthesis route, protein site identification, patients reaction and molecules interaction inside the drug, bringing us closer to the promise of precision medicine.

## 5. Application in hyperuricemia, gout arthritis, and hyperuricemic nephropathy
*5.1. Overview of hyperuricemia, gout arthritis, and hyperuricemic nephropathy*
Hyperuricemia can be characterized by an abnormally high blood uric acid level [177]. Generally, uric acid is produced from purine metabolism catalyzed mainly by XOD and can be oxidized to allantoin (a more soluble and easily excreted form) catalyzed mainly by uricase, which is crucial for uric acid level control in many animals including humans, but humans and other primates lack the uricase [178]. Uric acid and related metabolites are excreted through the kidney and gut [179]. Abnormal uric acid homeostasis including increased production, decreased degradation, and reduced excretion may ultimately result in hyperuricemia, which in turn leads to various complications, with gout arthritis, hyperuricemic nephropathy, diabetes mellitus, metabolic syndrome, and cardiovascular diseases as the most severe ones [180-183].

Driven by factors like aging populations, dietary changes, and comorbidities including obesity, metabolic syndrome, and chronic kidney diseases, the prevalence and incidence of hyperuricemia is rapidly rising worldwide, attracting extensive attention [184-188]. As a result, the incidence and prevalence of gout arthritis and hyperuricemic nephropathy are also increasing globally [189-191]. Over the past 20 years, the global incidence of gout has increased by 63.44%, and the global years lived with disability have increased by 51.12% [192]. During 1990-2019, the pace of increase in age-standardized rates of gout incidence, prevalence, and disability-adjusted life years accelerated with estimated annual percentage changes of 0.94 (95% CI: 0.85-1.03), 0.77 (95% CI: 0.69-0.84), and 0.93 (95% CI: 0.84-1.02), respectively [193]. In 2020, total number of gout prevalent cases globally was 558 (95% CI: 444-698) million, which was estimated to reach 958 (95% CI: 811-116) million in 2050, with an age-standardized prevalence of 667 (95% CI: 531-830) per 100 000 population and an increase rate of lager than 70% [194]. Taken together, these findings revealed a drastic rising trend of the global burden of disease, with urgent need for targeted public health interventions.

Unmet medication needs have been recognized for the treatment of hyperuricemia, gout arthritis, and hyperuricemic nephropathy. Currently, besides reducing purine intake (e.g., dietary control), therapies for hyperuricemia include XOD inhibitors (e.g., allopurinol, febuxostat, and topiroxostat), URAT1 inhibitors (e.g., probenecid and dotinurad), and recombinant uricases (e.g., pegloticase), as summarized in **Table 1** [195-198]. Colchicine, NSAIDs (e.g., naproxen and indomethacin), and steroids (e.g., prednisone) are available for the treatment of acute gout flares, but not chronic hyperuricemia, while colchicine and NSAIDs at low-doses can be used for the prevention of gout flares during initiation of urate-lowering therapies [199-201]. The IL1β monoclonal antibody (e.g., canakinumab) has also been newly approved for colchicine/NSAID-resistant flares. Those approved drugs for hyperuricemia and gout arthritis as mentioned above failed to offer high enough efficiency [202-204]. These drugs may as well exhibit significant side effects, such as CVD, SCAR, immunosuppression, myelosuppression, osteoporosis, hepatic/renal/gastrointestinal toxicity, etc., not to say that lesinurad (URAT1 inhibitors) and benzbromarone (XOD/URAT1 dual inhibitors) have been withdrawn due to serious adverse events [205-210]. Of note, hyperuricemic nephropathy is now managed off-label with allopurinol and febuxostat, since there are no drugs approved by FDA specifically for this indication [211]. All these indicate the big challenges in clinical practice and the urgent need for novel therapies with favorable efficacy and safety. Nevertheless, verinurad (URAT1 inhibitors), arhalofenate (URAT1 inhibitors/PPARγ agonists), SEL-212 (pegsiticase plus sirolimus nanoparticle), and uricase mRNA (gene therapy) are on the way to clinics.

Table 1 Approved and emerging drugs for hyperuricemia and gout arthritis.

| Drug | Status | Effect | Risk |
| --- | --- | --- | --- |
| Allopurinol | Approved, 1966, FDA | XOD inhibitors, inhibit uric acid synthesis | SCAR, rash, nausea, hepatic toxicity |
| Febuxostat | Approved, 2009, FDA | XOD inhibitors, inhibit uric acid synthesis | CVD, nausea, liver dysfunction |
| Topiroxostat | Approved, 2013, Japan | XOD inhibitors, inhibit uric acid synthesis | Liver dysfunction, GI discomfort |
| Probenecid | Approved, 1951, FDA | URAT1 inhibitors, block uric acid reabsorption | GI discomfort, rash, nephrolithiasis |
| Lesinurad | Withdrawn, 2019, FDA | URAT1 inhibitors, block uric acid reabsorption | Renal toxicity |
| Dotinurad | Approved, 2020, Japan | URAT1 inhibitors, block uric acid reabsorption | Renal impairment |
| Verinurad | Phase III | URAT1 inhibitors, block uric acid reabsorption | Renal adverse event |
| Benzbromarone | Withdrawn, 2019, Europe | XOD inhibitors, URAT1 inhibitors, dual actions | Hepatic toxicity, GI discomfort |
| Arhalofenate | Phase III | URAT1 inhibitors, PPARγ agonists, dual actions | Unclear |
| Pegloticase | Approved, 2010, FDA | Recombinant uricase, catalyze uric acid oxidation | Infusion reaction, gout flares, anaphylaxis |
| SEL-212 | Phase III | Pegsiticase plus sirolimus nanoparticle | Unclear |
| Uricase mRNA | Preclinical | mRNA-based uricase delivery | Unclear |
| Colchicine | Approved, 2009, FDA | Adjunct therapy, inhibit neutrophil activation | Myelosuppression, GI toxicity |
| Naproxen | Approved, 1994, FDA | Adjunct therapy, NSAIDs, anti-inflammatory, analgesic | GI ulcers, renal toxicity, hypertension |
| Indomethacin | Approved, | Adjunct therapy, NSAIDs, | GI ulcers, renal toxicity, |

| | 2010, FDA | anti-inflammatory, analgesic | hypertension |
|---|---|---|---|
| Prednisone | Approved, 2012, FDA | Adjunct therapy, steroids, anti-inflammatory | Immunosuppression, osteoporosis |
| Canakinumab | Approved, 2023, FDA | IL-1β monoclonal antibody, anti-inflammatory | Infection |

*5.2. Targets of hyperuricemia, gout arthritis, and hyperuricemic nephropathy*

The AI/ML techniques, along with conventional approaches, have been emergingly involved in target identification for hyperuricemia and related diseases. Network pharmacology, an approach widely applied in the field of target identification for drug discovery, is a hot spot for various human diseases based on natural products such as Chinese medicines in recent years, though proposed nearly two decades ago [212]. With the advancement in interdisciplinary disciplines including AI/ML and systematic biology, network pharmacology has been developed into a classic approach that hopefully improves the success rates of new drug clinical trials and saves the costs of new drug research and development. Many studies have applied network pharmacology along with relevant techniques to identify and validate potential targets of hyperuricemia, gout arthritis, and hyperuricemic nephropathy, as presented in **Table 2** [213-238], **Table 3** [239-264], and **Table 4** [265-285], respectively, generating many disease targets for further research and development in the future.

**Table 2** Identified targets of potential therapies for the management of hyperuricemia using network pharmacology and related techniques.

| Therapy | Target | Approach | Ref. |
|---|---|---|---|
| *Ampelopsis grossedentata* | ABCG2, PTGS2, XDH | ①②⑦⑨ | [213] |
| *Cardamine circaeoides* | ADA, AGTR1, JUN, PNP, PYGB, PYGL, PYGM, REN | ①④⑨ | [214] |
| Cortex Phellodendri Chinensis | ABCG2, AKT1, IL6, JUN, MAPK3, MAPK8, TP53, XDH | ①⑦⑨ | [215] |
| Dioscin | CAT, DDC, MAOA | ①②④⑨ | [216] |
| Folium Perillae | XDH | ①②③④ | [217] |
| Fructus Alpiniae Oxyphyllae | ESR1, HMGCR, PPARG, PTGS2 | ①②③⑦ | [218] |
| Fructus Gardeniae-Poria | AKT1, IL6, MAPK1, MYC, VEGFA | ①②⑨ | [219] |
| Fructus Lycii | ALDH2, APOB, ESR1, HTR2A, IL6, MAOA, MMP3, NOS3, PPARG, PTGS2, TNF, VEGFA, XDH | ①②③④⑦⑨ | [220] |
| Fufang Qiling Granule | XDH | ①②⑨ | [221] |
| Gegen Qinlian Decoction | HDAC3, NLRP3, NR1D1, RORA | ①②③⑤⑩ | [222] |
| Hazel leaf polyphenols | BRCA1, CASP3, ESR1, IL6, MDM2, MYC, NFKB1, PPARG, VEGFA | ①②⑨ | [223] |
| Herba Cichorii Radix Cichorii | CASP3, ESR1, PTGS2, RELA, TNF | ①②⑨ | [224] |
| Herba Plantaginis | AKT1, AR, ESR1, MAPK1, MAPK3, PTGS2, SIRT1, TNF | ① | [225] |
| Herba Portulacae | ABCG2, SLC2A9, SLC22A12, XDH | ①②⑦⑨ | [226] |
| Isobavachin | ABCG2, HPRT1, REN | ①② | [227] |
| Paidu Powder | CASP3, ESR1, IL1B, IL6, MAPK1, MAPK8, RELA, TNF, TP53 | ①②⑩ | [228] |
| Pericarpium Zanthoxyli | ESR1, MAOA, PTGS2, TNF, XDH | ①②③ | [229] |
| Poria | FASN, XDH | ①②④⑨ | [230] |
| *Poria cocos* solid fermenting Radix Astragali | CASP3, ESR1, JUN, STAT3, TNF | ①②⑦⑨ | [231] |
| Qushi Huazhuo Tongluo Decoction | CASP3, CAT, CCL2, IL1B, IL6, IL10, MAPK3, PPARG, TNF, VEGFA | ①②⑨ | [232] |
| Radix ginseng-Fructus Jujubae | IL1B, TNF, VEGFA | ①②⑨ | [233] |
| Saffron petal | ABCG2, SLC2A9, SLC22A12 | ①②⑨ | [234] |
| Shizhifang Decoction | CASP3, ESR1, GAPDH, IL1B, IL6, INS, MYC, PPARG, PTGS2, TNF, TP53, VEGFA | ①②⑦⑨ | [235] |
| Simiao Pill | ABCG2, CASP3, INS, PTGS2, RELA, SLC2A9, SLC22A6, SLC22A12, TLR4, XDH | ①②⑨ | [236] |
| Wuling Powder | MAPK1, RELA, TNF, TP53 | ①② | [237] |
| Wuling Powder | CASP3, ESR1, HSP90AA1, IL6, MAPK3, MTOR, NOS3, PPARG, PTGS2, SIRT1, STAT3, TNF | ①②⑨ | [238] |

**Notes**: ① = network pharmacology; ② = molecular docking; ③ = molecular dynamics simulation; ④ = metabolomics; ⑤ = transcriptomics; ⑥ = metagenomics; ⑦ = in vitro; ⑧ = in situ; ⑨ = in vivo; ⑩ = clinical study.

Table 3 Identified targets of potential therapies for the management of gout arthritis using network pharmacology and related techniques.

| Therapy | Target | Approach | Ref. |
|---|---|---|---|
| Baicalin | MMP9, PTGS2, TLR4, TNF, VEGFA | ①②⑦ | [239] |
| Biqi Capsule | ABCB1, CREBBP, EP300, HSP90AA1, ITGB1, MAPK8, PTGS2, RELA, SRC, STAT3 | ①②⑦⑨ | [240] |
| Caulis Sinomenii | HIF1A, JUN, MMP9, NOS2, PPARG, PTGS2, SRC, STAT3, TLR4 | ①② | [241] |
| Celery seed | AKT1, EGFR, HSP90AA1, ITGB1, MAPK1, MAPK3, MAPK8, MAPK14, NOS2, PIK3R1, PIK3CA, PPARA, RELA | ①②③ | [242] |
| Danggui Niantong Decoction | CASP8, CXCL8, FOS, IL1B, IL6, JUN, MMP1, PTGS2, STAT1, TNF | ①② | [243] |
| *Dioscorea septemloba* | ESR1, HIF1A, HSP90AA1, MMP9, MTOR, PPARG, PTGS2, STAT3, TLR4 | ①② | [244] |
| Ermiao Powder | AKT1, CASP3, EGF, EGFR, FN1, HIF1A, IL1B, IL6, JUN, MMP9, MYC, PTGS2, TP53, VEGFA | ①② | [245] |
| Folium Mori | AKT1, PLA2G2A, PRKCA | ①② | [246] |
| Fructus Piperis Longi | AKT1, MAPK, PIK3CA | ①②⑨ | [247] |
| Guizhi Shaoyao Zhimu Decoction | CASP3, CXCL8, ESR1, PTGS2 | ① | [248] |
| Pachyman | ACE, APP, HSP90AA1, IL2, MAPK14, MMP1, MMP2, NOS3, SRC, STAT3, TNF | ①②⑨ | [249] |
| *Paederia foetida* | BCL2, HSP90AA1, IL1B, IL6, MAPK1, MAPK8, PTGS2, TNF, TP53 | ①② | [250] |
| Polyporus | IL6, MMP9, PPARG, PTGS2, SRC, TNF | ①②③⑦ | [251] |
| Poria cum Radix Pini | ALB, IL1B, PPARA, PPARG, PTGS2, REN, TNF | ①②⑦ | [252] |
| Qianhua Gout Capsule | CXCL8, IL1B, IL6, MMP9, PPARG | ①②⑨ | [253] |

| Therapy | Target | Approach | Ref. |
|---|---|---|---|
| Quercetin | CCL2, IL1A, IL1B, IL2, IL6, IL10, MMP1, MMP9, TNF, VEGFA | ①⑨ | [254] |
| Quzhuo Tongbi Decoction | AKT1, CASP3, IL6, JUN, MMP9, PTGS2, TNF, TP53, VEGFA | ①②④⑨ | [255] |
| Radix Gentianae Macrophyllae | EGFR, ERBB2, GAPDH, IL2, MAPK3, PPARA, PPARG, PTGS2, PTPRC, STAT3 | ①⑨ | [256] |
| Simiao Pill | APP, CXCL8, CXCR2, GCK, IGF1R, INSR, JAK2, PIK3CA, PIK3R1, PPARG, PTGS2, REN, SIRT1, STAT3, TLR4 | ①② | [257] |
| Simiao Powder | AHR, IL6, NFKBIA, PPARG, PRKCA, PTGS1, VEGFA | ①②⑨ | [258] |
| Tongfengding Capsule | AKR1C3, ALOX5, CYP2B6, PLA2G1B, PTGS1, PTGS2 | ①⑦⑧⑨ | [259] |
| Tongfengting Decoction | AKT1, IL6, TNF, TP53, VEGFA | ①② | [260] |
| Wuwei Shexiang Pill | CASP1, NLRP3, RELA, TLR4 | ①②⑦⑨ | [261] |
| Wuwei Xiaodu Drink | APOE, IL1β, IL6, IL10, INS, LEP, TLR4, TNF | ①⑨ ⑩ | [262] |
| Yinhua Gout Granule | IL1B, IL6, NOS2, PTGS1, PTGS2, TNF | ①②⑤⑨ | [263] |
| Zisheng Shenqi Decoction | IL1B, NLRP3, P2RX7 | ① | [264] |

**Notes**: ① = network pharmacology; ② = molecular docking; ③ = molecular dynamics simulation; ④ = metabolomics; ⑤ = transcriptomics; ⑥ = metagenomics; ⑦ = in vitro; ⑧ = in situ; ⑨ = in vivo; ⑩ = clinical study.

**Table 4** Identified targets of potential therapies for the management of hyperuricemic nephropathy using network pharmacology and related techniques.

| Therapy | Target | Approach | Ref. |
|---|---|---|---|
| Baicalin and baicalein | AKT1, CAT, ENO2, ITGB1, MMP3, NFE2L2, PTGS1, REN, SOD1, STAT3, TP53, XDH | ①②⑨ | [265] |
| Berberine | AKT1, ESR1, IGF1, IL1B, IL6, INS, JUN, MAPK1, MAPK3, MAPK8, MYC, STAT3, TNF, TP53, VEGFA | ①②⑨ | [266] |
| Cheqianzi | ALB, CASP3, IL1B, IL6, INS, JUN, MAPK3, | ①⑨ | [267] |
| Decoction | PPARG, STAT3, TNF, TP53 | | |
| Cinnamon essential oil | IL1B, NLRP3, P2RX7, PTGS2, XDH | ①②④⑨ | [268] |
| *Clerodendranthus spicatus* aqueous extract | EP300, GRB2, HRAS, MAPK1, MAPK3, PIK3CA, PIK3R1, RELA, SRC, STAT3 | ①②⑨ | [269] |
| Fructus Chebulae | ALB, CASP3, EGFR, GAPDH, IL6, MAPK3, SRC, STAT3, TNF | ①⑦⑨ | [270] |
| Fructus Phyllanthi | ABCG2, IRS1, PTGS2, SLC22A6, SLC22A12, TLR4, XDH | ①②③⑦⑨ | [271] |
| Fuling Zexie Formula | APP, CASP3, ESR1, HSP90AA1, IL6, MAPK1, PPARA, PPARG, PTGS2, STAT3, TNF | ①⑨ | [272] |
| Galangin | ALB, CASP3, HSP90AA1, STAT3, TNF, TP53 | ①②④⑨ | [273] |
| Herba Taxilli | ALB, BCL2, GAPDH, IL6, JUN, MYC, NFKB1, PPARG, TNF, TP53 | ①②⑦⑨ | [274] |
| Isorhamnetin | CDK6, IGF1R, KDR, MCL1, PARP1, PIK3CG, PTGS2, RELA, SYK | ①④⑨ | [275] |
| Jiangniaosuan Formula | BCL2, CASP3, HIF1A | ①④⑦⑨ | [276] |
| Juju Formula | APP, IL6, JUN, MAPK1, MAPK3, MAPK8, RELA, STAT3, TNF, TP53 | ①②⑥⑨ | [277] |
| Juju Formula | CASP3, IL6, JUN, TNF, TP53, VEGFA | ①②⑨ | [278] |
| *Lavandula latifolia* essential oil | ACE, AMPK1, ESR1, HSP90AA1, MAPK3, NFKB1, PPARG, PTGS2, TNF | ①②④⑤⑦⑨ | [279] |
| *Phellinus igniarius* | ALDH2, APOB, CAT, CXCL8, FASN, GPT, PC, PPARG, UGT1A1, XDH | ①⑨ | [280] |
| Simiao Pill | BCL2L1, BIRC3, FOS, IL1B, NFKB1, TGFB3 | ①②⑤④⑨ | [281] |
| Sinapic acid | CTNNB1, EGFR, ESR1, IL1B, MAPK8, STAT3, TNF | ①②⑦⑨ | [282] |
| β-Sitosterol | BCL2, CASP3, CASP9, ESR1, HMGCR, IGF1R, MDM2, NFE2L2 | ①④⑨ | [283] |
| Vine grape tea polyphenols | BAX, BCL2, CASP3, STAT3, TGFB1, TP53, XDH | ①②⑦⑨ | [284] |
| *Xanthoceras sorbifolium* leaves | AKT1, ALB, CASP3, EGFR, HRAS, HSP90AA1, IGF1, MMP9, SRC | ①⑦⑨ | [285] |

**Notes**: ① = network pharmacology; ② = molecular docking; ③ = molecular dynamics simulation; ④ = metabolomics; ⑤ = transcriptomics; ⑥ = metagenomics; ⑦ = in vitro; ⑧ = in situ; ⑨ = in vivo; ⑩ = clinical study.

According to the findings, some unique targets could be highlighted as potential candidates. For purine and uric acid metabolism (**Figure 3**), along with the well-established XOD (XOD can be converted from XDH) that catalyzes hypoxanthine into xanthine and then uric acid, ADA, PNP, and HPRT1 (HGPRT) have also been recognized as key enzymes. As demonstrated, ADA catalyzes adenosine into inosine, PNP reversibly catalyzes inosine into hypoxanthine and guanosine into guanine, while HPRT1 catalyzes hypoxanthine into inosine monophosphate and guanine into guanosine monophosphate, thus ADA inhibitors, PNP inhibitors, and HPRT1 agonists may have the potential as candidates to reduce uric acid production [214, 227]. For uric acid and urate excretion (**Figure 4**), besides the highly specific SLC22A12 (URAT1) that is responsible for 90% urate reabsorption, SLC2A9 (GLUT9), SLC22A6 (OAT1), and ABCG2 have also been demonstrated as important transporters. In brief, SLC2A9 is an uptake transporter, while SLC22A6 and ABCG2 are export transporters, thence SLC2A9 inhibitors, SLC22A6 agonists, and ABCG2 agonists may serve as alternative uricosurics to increase uric acid excretion [226, 234, 236, 271]. For inflammasome-related inflammatory response (**Figure 5**), in addition to IL1B with newly approved monoclonal antibody canakinumab, antibodies or inhibitors targeting P2RX7, NLRP3, and CASP1 may also deserve further exploration, as P2RX7 may mediate the activation of NLRP3 inflammasome, which facilitates the generation of active CASP1 that in turn processes pro-IL-1β and pro-IL-18 to the active mature forms [261, 264, 268]. For kidney disorder without specific drugs, since the MSU crystal may not only induce inflammation but also apoptosis in kidney residential cells, targeting BAX, BCL2, BRCA1, CASP3, CASP8, MCL1, and PARP1 involved in the extrinsic and intrinsic apoptotic pathways (**Figure 6**) may also be taken into consideration [223, 243, 275, 284]. Of note, further translational and clinical studies involving humans are warranted to further confirm the exact pathophysiological roles of the targets in disease course and the feasibility of corresponding drugs to produce desired therapeutic effects.

In contrast to the above presentations, some distinct findings are of clinical significance as well. By combining association rule learning with gene network analysis, a *puuD* gene-encoded integral membrane protein with a C-terminal cytochrome *c* was identified as the missing urate oxidase with a domain of unknown function (DUF989) responsible for the biochemical activity, completing the link in the conversion of urate into allantoin and pointing out an enzyme useful for drug development [286]. Using ML, chemoinformatic analysis, and integrative systems metabolic analysis of transcriptomic and metabolomic data, it was revealed that loss of kidney URAT1 led to disrupted redox homeostasis regarding reactive oxygen species handling, vitamin C metabolism, and cofactor charging reactions, implicating the need to thinking twice about the side effects of URAT1 inhibition when using URAT1 inhibitors for hyperuricemia management [287]. Through ML algorithms and transcriptomic analysis, five purine metabolism-related genes (ADK, DCK, GMPS, NME5, and POLR3D) were linked to immune reprogramming concerning fluctuation in the proportion and activity of various immunocytes as well as the survival of tumor patients, informing the risk of developing immune disorders and tumors in gout patients as well as the possibility of treating these complications with concurrent targets [288].

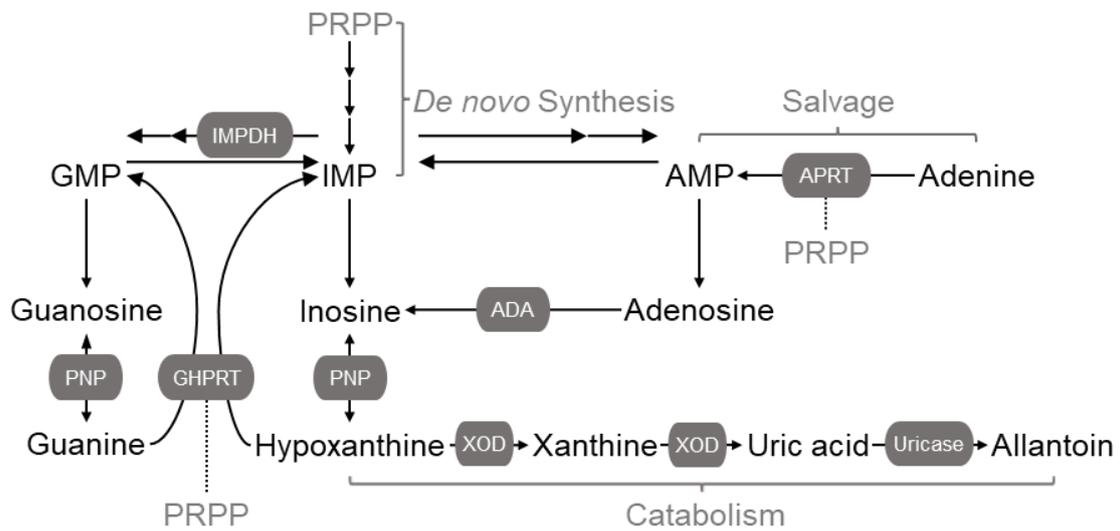

Figure 3 Uric acid metabolism pathway.

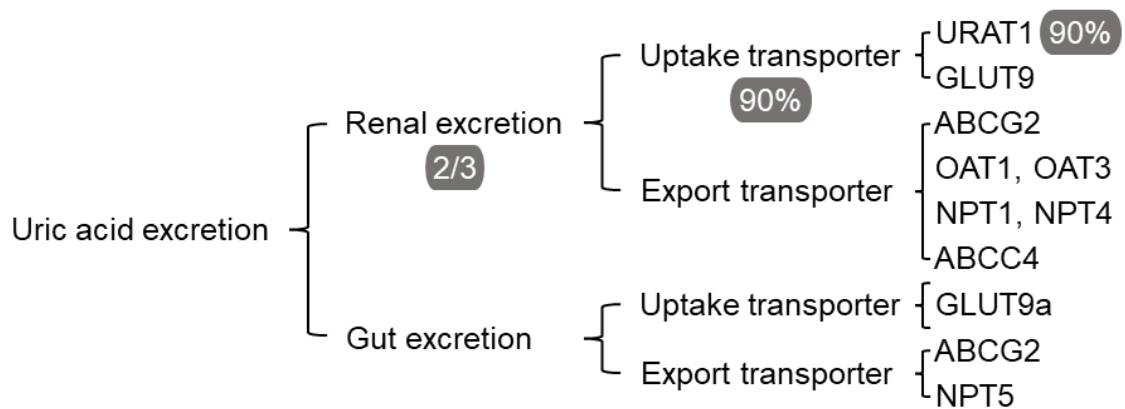

Figure 4 Uric acid excretion pathway.

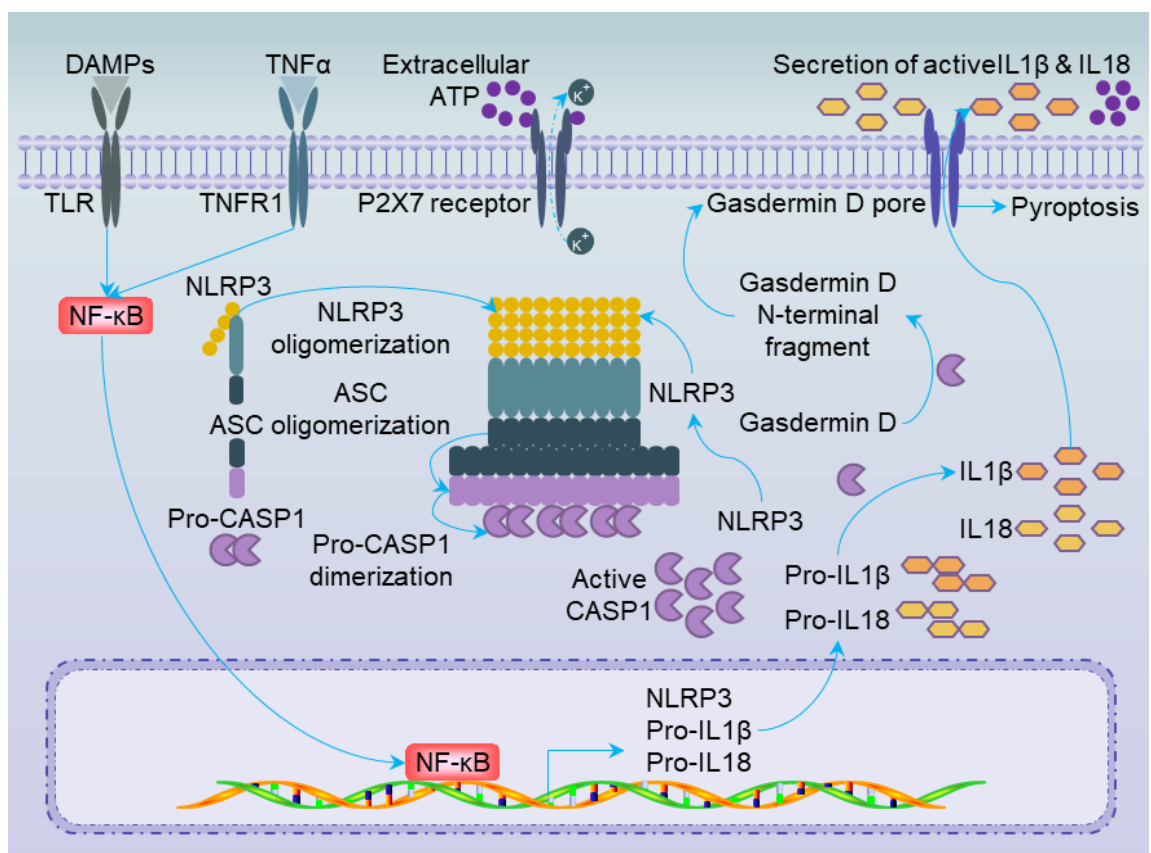

**Figure 5** NLRP3 inflammasome signaling pathway.

*5.3. Hits of hyperuricemia, gout arthritis, and hyperuricemic nephropathy*

Great efforts have been made to screen hit compounds for hyperuricemia, taking advantage of novel AI/ML-based approaches and related techniques. Some studies have particularly focused on the screening of XOD inhibitors. After creating a molecular database of natural XOD inhibitors, developing QSAR models, and conducting experimental validation and molecular docking, it was reported that the hydrophobicity and steric molecular structures may be the features of XOD inhibitors, and vanillic acid may serve as a promising new candidate of XOD inhibitors [289]. Using a similar hybrid approach combining ML-based QSAR modeling with molecular docking and validation by various metrics, compounds NK-1 and NK-2 out of alkannin/shikonin derivatives were confirmed to have pIC50 value equivalent to allopurinol [290]. With a consensual dataset of extended connectivity fingerprints (ECFPs) of 249 XOD inhibitors and ML-assisted QSAR model using with cross-validation/external validation, SHAP interpretation, molecular docking simulation, and ADMET prediction, 15 new molecules were deemed as superior XOD inhibitors, and molecules with selenazole moiety, cyano group, and isopropyl group were proposed to possess potent XOD inhibitory activities [291]. In another study involving a comprehensive database of 315 XOD inhibitors and 28 ML-based QSAR models along with molecular docking, molecular dynamics simulations, ADME predictions, and in vitro assay validation, daphnetin, 7-hydroxycoumarin, and piceatannol were identified as potential XOD inhibitors possessing desirable drug-like properties and safety characteristics, with piceatannol showing a remarkable stability through hydrogen bonding and hydrophobic interactions as well as the strongest inhibitory activity [292]. These studies have demonstrated a strategy for efficient identification of natural XOD inhibitors, accurate prediction of inhibitory potency of XOD inhibitors, and remarkable explication of interaction mechanisms between XOD inhibitors and XOD.

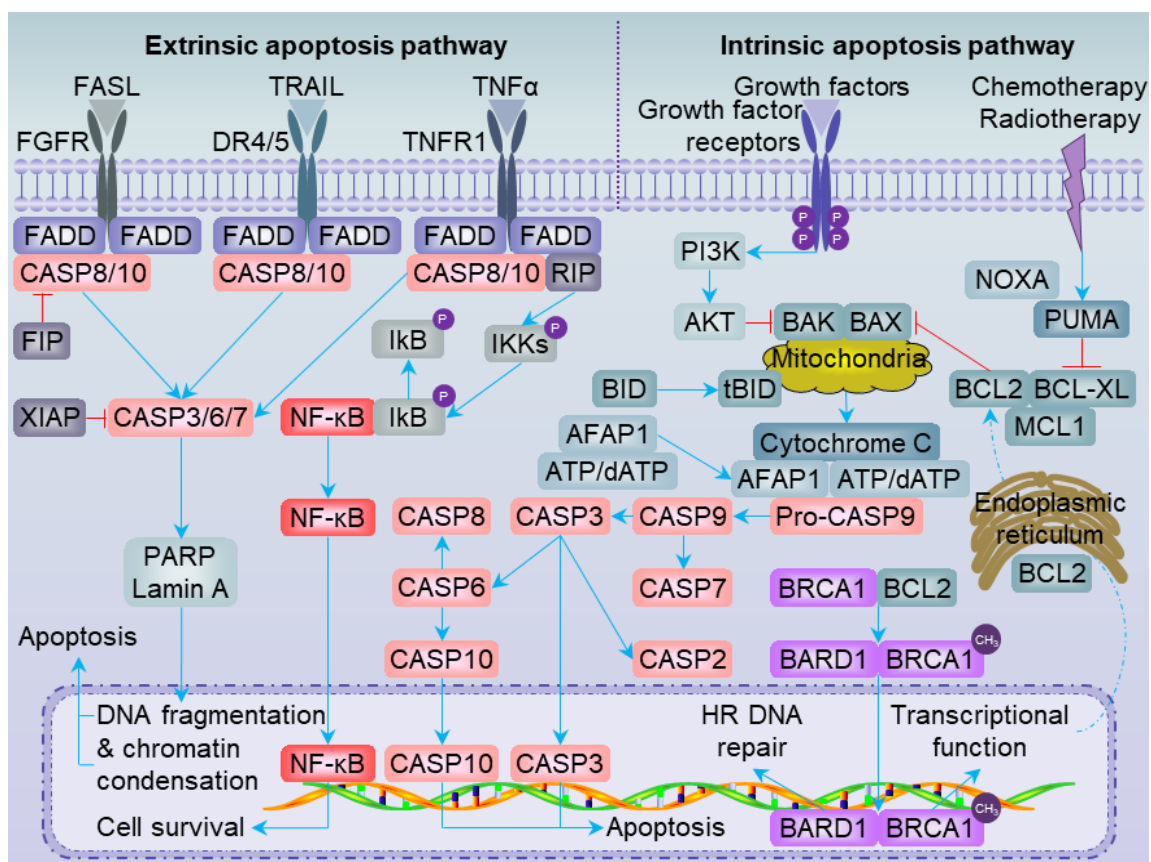

**Figure 6** Cell apoptosis signaling pathway.

In addition to conventional XOD inhibitors from small molecules, AI/ML-based screening approaches have also been applied to identify peptides with XOD inhibitory properties. Through virtual-screening, ML-based predicting, and molecular-docking, the tripeptide, tetrapeptide, pentapeptide, and hexapeptide were identified as XOD inhibitory peptides, with residues W, F, and G as the key amino acids in the binding sites with XOD [293]. Using DL in combination with molecular docking and molecular dynamics simulations, two food-derived peptides, LWM and ALPM, were unraveled as noncompetitive inhibitors of XOD, with LWM showing a thinner, longer, and more twisted configuration compared to ALPM [294]. With similar approaches combining ML-based screening, molecular docking, molecular dynamics simulations and experimental validations, two peptides ECFK and FECK from the *Bacillus subtilis* proteome were identified and validated as XOD inhibitors with sensory properties and favorable safeties, and ECFK showed a higher dissociation energy barrier than FECK (indicating strong and stable interactions) [295]. These works provided novel insights into the screening of XOD inhibitors from peptides, highlighting food-derived peptides as potent and safe natural inhibitors beyond other small molecules.

As for various diseases or disorders without obvious targets to enable drug discovery based on target proteins, novel approaches may be warranted to overcome the challenges behind. Notably, a DL-based efficacy prediction system (DLEPS) has been described, which may identify drug candidates using a change in the gene expression profile in the diseased state as input. With prediction by DLEPS and validation by experiments, perillen was successfully identified with XOD inhibitory property as well as significant effects against hyperuricemia, inflammation, and renal fibrosis in mice [296]. Success cases of DLEPS have also been observed in identifying chikusetsusaponin IV and trametinib for obesity and nonalcoholic steatohepatitis, respectively. DLEPS may serve as an effective tool for drug repurposing and discovery, suggesting drug discovery based on ML-driven transcriptional profiling may not necessarily require data of protein targets, which may reform drug discovery process.

*5.4. Trials of hyperuricemia, gout arthritis, and hyperuricemic nephropathy*
After target identification and validation as well as hit screening and validation, the initial compounds with desired activity are then optimized to improve their suitability and drug-like properties, such as bioactivity, selectivity, PK, pharmacodynamics, and toxicity, ultimately fielding the drug candidates with the best chance of success in clinical trials. Confronted with the high-risk context of increasing cost, duration, and failure rate, clinical trial optimization is another important part of drug development program to answer important research questions more efficiently and with less risk, via improving dose selection, patient stratification, and trial design. Upon systematic literature searching and screening, there are no sufficient AI/ML-driven approaches specifically designed for lead optimization in drug discovery and development for hyperuricemia and related diseases. Yet one platform established with a comprehensive database of 315 XOD inhibitors, 28 ML-based QSAR models, molecular docking, molecular dynamics simulations, ADME predictions, and in vitro assay validation may also be applied partially for hi-to-lead optimization of XOD inhibitors [292]. In addition, AI/ML-driven techniques are also revolutionizing the healthcare of hyperuricemia, gout arthritis, and hyperuricemic nephropathy by facilitating risk prediction, disease diagnosis, disease classification, precision medicine, and prognosis prediction [297-310]. These studies may provide reference for better clinical trial design in the late stages of drug discovery for hyperuricemia, gout arthritis, and hyperuricemic nephropathy.

## 6. Discussion and perspective

In the previous sections, we examined how AI/ML techniques have transformed each stage of the drug discovery process. Here, we highlight some general limitations and challenges that AI/ML techniques currently face within drug discovery and outline several promising future directions from both computational and biological perspectives.

One of the primary hurdles in AI-driven drug discovery is the limited availability and uneven quality of training data. Many current datasets are small, domain-specific, and suffer from selection bias and class imbalance issues. For instance, kinase inhibitor datasets are often well-documented, whereas many other protein families or therapeutic targets remain significantly underrepresented. Additionally, the lack of standardized data formats, experimental methods, and annotations limits model generalizability and consistency across studies. Moreover, integrating multimodal datasets—such as omics, EHRs, and real-time biosensor data—continues to be challenging due to interoperability issues and data privacy concerns [311, 312].

AI and ML models, particularly DL systems, often function as "black boxes," yielding accurate predictions but providing limited transparency regarding their decision-making processes. This opacity reduces their credibility in clinical and regulatory settings, posing significant barriers for approval by authorities such as the FDA or EMA. Although explainability methods like SHAP, LIME, and attention visualizations in transformer models provide some insights, their outputs can still be difficult for domain specialists to interpret comprehensively. Enhanced interpretability is especially critical in applications like toxicity prediction and PK modeling, where understanding causal relationships and biological mechanisms is more valuable than prediction accuracy alone [22, 313-315].

Many AI-driven methods are validated primarily through computational (in silico) means, which often fail to translate accurately into experimental (in vitro) or biological (in vivo) contexts. Discrepancies frequently arise due to the complexity of biological systems, individual patient variability, and experimental uncertainty. Additionally, the absence of standardized benchmarking frameworks and gold-standard datasets makes it challenging to compare algorithms and validate performance across different studies. Although initiatives like the Tox21 challenge have advanced benchmarking for toxicity prediction, similar efforts are necessary for evaluating models related to ADME, clinical trial simulation, and digital twin applications [316, 317].

Emerging computational technologies such as quantum ML show great promise for overcoming current limitations in molecular modeling and drug discovery. Quantum ML has the potential to significantly speed up conformational sampling, electronic property computations, and the modeling of complex protein-ligand interactions. Simultaneously, the rise of digital twin frameworks—virtual patient models constructed using real-world EHR data, genetic information, and lifestyle factors—may enable highly personalized simulations of drug responses, greatly improving clinical trial design, safety assessments, and pharmacodynamics predictions. These computational advancements also offer the potential to reduce reliance on animal models, thus providing more accurate, human-specific insights [318, 319].

AI's most transformative potential in drug discovery may lie in its ability to complement human expertise rather than replace it. Leading research institutions are already adopting robotic laboratories that integrate automated chemical synthesis platforms with AI-driven retrosynthesis planners and property optimization engines. Moreover, federated learning frameworks, which enable collaboration across institutions without requiring direct sharing of raw data, can help mitigate data scarcity and privacy issues. Such approaches are particularly beneficial in the context of rare disease research and low-resource environments, where individual institutions typically have limited datasets [320, 321].

To fully realize the potential of AI-driven drug discovery, regulatory frameworks and ethical standards must evolve in parallel. Agencies like the FDA are already exploring adaptive regulatory mechanisms that could accommodate AI-designed pharmaceuticals and digital biomarkers. Furthermore, addressing ethical challenges—including algorithmic bias, equitable data ownership, and intellectual property rights for AI-generated molecular entities—is essential to prevent future legal and societal conflicts. The establishment of industry-wide standards for model documentation, validation procedures, and interpretability measures will be critical for ensuring clinical acceptance and commercialization of AI-driven solutions [322].

## 7. Conclusion

AI/ML have emerged as transformative forces capable of reshaping most the stages of the drug discovery pipeline—from target identification to hit screening and lead optimization that hopefully provide significant foundations to promote success in the later clinical trials. By harnessing extensive biological, chemical, and clinical datasets, the AI/ML methodologies have significantly enhanced the efficiency, accuracy, and cost-effectiveness of drug discovery. The detailed assessment of AI/ML techniques presented in this review, reinforced by a comprehensive case study on hyperuricemia, gout arthritis, and hyperuricemic nephropathy, underscores both the substantial successes and ongoing challenges faced by AI/ML integration within the fields of biomedical research and pharmaceutical industry.

Despite promising advancements, AI/ML-driven drug discovery must still overcome critical limitations, such as data scarcity, insufficient model interpretability, and challenges related to experimental validation and benchmarking. Addressing these barriers will require interdisciplinary collaboration among computational scientists, biologists, clinicians, and regulatory bodies. Promising future research directions, including quantum ML, digital twin frameworks, and federated learning, as well as applications in the discovery of innovative macromolecule therapies from proteins, peptides, polymers, and nucleic acids distinguished from traditional small molecules, provide viable pathways to surmount current limitations and accelerate the adoption of AI/ML technologies in drug discovery.

Ultimately, realizing the full potential of AI/ML in drug discovery and beyond other biomedical research necessitates not only technological innovation but also concurrent regulatory evolution and careful consideration of ethical implications. Continued refinement of computational methodologies, establishment of standardized validation practices, and enhancement of model transparency will enable AI/ML-driven drug discovery to fulfill its promise—accelerating therapeutic innovation and significantly improving healthcare outcomes.

**Acknowledgement**
This work was supported by Health Bureau-Health and Medical Research Fund (21222101, 18192141), Health Bureau-Chinese Medicine Development Fund (22B2/007A, 19SB2/002A), Research Grants Council, University Grants Committee-General Research Fund (17121419), Innovation and Technology Commission-Innovation and Technology Fund-Partnership Research Programme (PRP/069/22FX).

**CRediT authorship contribution statement**
**Junwei Su**: Conceptualization, Writing - Original Draft, Writing - Review & Editing, Validation. **Cheng Xin**: Conceptualization, Writing - Original Draft, Writing - Review & Editing, Validation. **Ao Shang**: Writing - Original Draft, Validation. **Shan Wu**: Writing - Review & Editing, Validation. **Zhenzhen Xie**: Writing - Review & Editing, Validation. **Ruogu Xiong**: Writing - Review & Editing, Validation. **Xiaoyu Xu**: Writing - Review & Editing, Validation. **Cheng Zhang**: Writing - Review & Editing, Validation. **Guang Chen**: Writing - Review & Editing, Validation. **Yau-Tuen Chan**: Writing - Review & Editing, Validation. **Guoyi Tang**: Conceptualization, Writing - Original Draft, Writing - Review

& Editing, Validation, Project Administration. **Ning Wang**: Writing - Review & Editing, Validation, Project Administration. **Yong Xu**: Writing - Review & Editing, Validation. **Yibin Feng**: Conceptualization, Writing - Review & Editing, Validation, Supervision, Funding Acquisition.

**Declaration of competing interest**

The authors declare that there is no conflict of interest.

**Data availability**

No data was used for the work described in this review article.

**Abbreviations**

ABCB1/G2, ATP binding cassette subfamily B member 1/subfamily G member 2; ACE, angiotensin I converting enzyme; ADA, adenosine deaminase; ADK, adenosine kinase; ADME, absorption, distribution, metabolism, and excretion; AGTR1, angiotensin II receptor type 1; AHR, aryl hydrocarbon receptor; AI, artificial intelligence; AKR1C3, aldo-keto reductase family 1 member C3; AKT1, AKT serine/threonine kinase 1; ALB, albumin; ALDH2, aldehyde dehydrogenase 2 family member; ALOX5, arachidonate 5-lipoxygenase; ANN, artificial neural network; APOB/O, apolipoprotein B/O; APP, amyloid beta precursor protein; AR, androgen receptor; BAX, BCL2 associated X, apoptosis regulator; BCL2, BCL2 apoptosis regulator; BCL2L1, BCL2 like 1; BIRC3, baculoviral IAP repeat containing 3; BRCA1, BRCA1 DNA repair associated; CASP1/3/8, caspase 1/3/8; CAT, catalase; CCL2, C-C motif chemokine ligand 2; CDK6, cyclin dependent kinase 6; CI, confidence interval; CNN, convolutional neural network; CREBBP, CREB binding lysine acetyltransferase; CTNNB1, catenin beta 1; CVD, cardiovascular death; CXCL8, C-X-C motif chemokine ligand 8; CXCR2, C-X-C motif chemokine receptor 2; CYP2B6, cytochrome P450 family 2 subfamily B member 6; DCK, deoxycytidine kinase; DDC, dopa decarboxylase; DL, deep learning; DNN, deep neural network; EGF, epidermal growth factor; EGFR, epidermal growth factor receptor; EHR, electronic health record; EMA, European Medicines Agency; ENO2, enolase 2; EP300, EP300 lysine acetyltransferase; ERBB2, erb-b2 receptor tyrosine kinase 2; ESR1, estrogen receptor 1; FASN, fatty acid synthase; FDA, U.S. Food and Drug Administration; FN1, fibronectin 1; FOS, Fos proto-oncogene, AP-1 transcription factor subunit; GAN, generative adversarial network; GAPDH, glyceraldehyde-3-phosphate dehydrogenase; GAT, graph attention network; GCK, glucokinase; GCN, graph convolutional network; GI, gastrointestinal; GLUT9, glucose Transporter 9; GMPS, guanine monophosphate synthase; GNN, graph neural network; GPT, glutamic--pyruvic transaminase; GRB2, growth factor receptor bound protein 2; HDAC3, histone deacetylase 3; HGPRT, hypoxanthine-guanine phosphoribosyltransferase; HIF1A, hypoxia inducible factor 1 subunit alpha; HMGCR, 3-hydroxy-3-methylglutaryl-CoA reductase; HPRT1, hypoxanthine phosphoribosyltransferase 1; HRAS, HRas proto-oncogene, GTPase; HSP90AA1, heat shock protein 90 alpha family class A member 1; HTR2A, 5-hydroxytryptamine receptor 2A; IGF1, insulin like growth factor 1; IGF1R, insulin like growth factor 1 receptor; IL1A/1B/2/6/10, interleukin 1 alpha/1 beta/2/6/10; INS, insulin; INSR, insulin receptor; ITGB1, integrin subunit beta 1; JAK2, Janus kinase 2; JUN, Jun proto-oncogene, AP-1 transcription factor subunit; KDR, kinase insert domain receptor; LBVS, ligand-based virtual screening; LEP, leptin; LLM, large language model; LIME, local interpretable model-agnostic explanations; MAOA, monoamine oxidase A; MAPK1/3/8/14, mitogen-activated protein kinase 1/3/8/14; MCL1, MCL1 apoptosis regulator, BCL2 family member; MDM2, MDM2 proto-oncogene; ML, machine learning; MMP1/2/3/9, matrix metallopeptidase 1/3/9; MTOR, mechanistic target of rapamycin kinase; MYC, MYC proto-oncogene, bHLH transcription factor; NFE2L2, NFE2 like bZIP transcription factor 2; NFKB1, nuclear factor kappa B subunit 1; NFKBIA, NFKB inhibitor alpha; NLP, natural language processing; NLRP3, NLR family pyrin domain containing 3; NME5, NME/NM23 family member 5; NOS2/3, nitric oxide synthase 2/3; NR1D1, nuclear receptor subfamily 1 group D member 1; NSAIDs, non-steroidal anti-inflammatory drugs; OAT1, organic anion transporter 1; P2RX7, purinergic receptor P2X 7; PARP1, poly(ADP-ribose) polymerase 1; PBPK, physiologically based pharmacokinetic; PC, pyruvate carboxylase; PIK3CA/G, phosphatidylinositol-4,5-bisphosphate 3-kinase catalytic subunit alpha/gamma; PIK3R1, phosphoinositide-3-kinase regulatory subunit 1; PK, pharmacokinetics; PLA2GIB/2A, phospholipase A2 group IB/IIA; PLM, protein language model; PNP, purine nucleoside phosphorylase; POLR3D, RNA polymerase III subunit D; PPARA/G, peroxisome proliferator activated receptor alpha/gamma; PPI, protein-protein interaction; PRKCA, protein kinase C alpha; PTGS1/2, prostaglandin-endoperoxide synthase 1/2; PTPRC, protein tyrosine phosphatase receptor type C; PYGB/L/M, glycogen phosphorylase B/L/muscle associated; QSAR, quantitative structure-activity relationship; RELA, RELA proto-oncogene, NF-kB subunit; REN, renin; RF, random forest; RL, reinforcement learning; RORA, RAR related orphan receptor A; SBVS, structure-based virtual screening; SCAR, severe cutaneous adverse reaction; SHAP, SHapley Additive Explanations; SIRT1, sirtuin 1; SLC2A9/22A6/22A12, solute carrier family 2 member 9/family 22 member 6/family 22 member 12; SOD1, superoxide dismutase 1; SRC, SRC proto-oncogene, non-receptor tyrosine kinase; STAT3, signal transducer and activator of transcription 3; SVM, support vector machine; SYK, spleen associated tyrosine kinase; TGFB3, transforming growth factor beta 3; TLR4, toll like receptor 4; TNF, tumor necrosis factor; TP53, tumor protein p53; UGT1A1, UDP glucuronosyltransferase family 1 member A1; URAT1, urate transporter 1; VAE, variational autoencoders; VEGFA, vascular endothelial growth factor A; XDH, xanthine dehydrogenase; XOD, xanthine oxidase.

Opinion on Drug Discovery, 2024. **19**(7): p. 841-853.

15. Vatansever, S., et al., Artificial intelligence and machine learning-aided drug discovery in central nervous system diseases: State-of-the-arts and future directions. Medicinal Research Reviews, 2021. **41**(3): p. 1427-1473.

16. Dara, S., et al., Machine Learning in Drug Discovery: A Review. Artificial Intelligence Review, 2022. **55**(3): p. 1947-1999.

17. Sadybekov, A.V. and V. Katritch, Computational approaches streamlining drug discovery. Nature, 2023. **616**(7958): p. 673-685.

18. Tropsha, A., et al., Integrating QSAR modelling and deep learning in drug discovery: the emergence of deep QSAR. Nature Reviews Drug Discovery, 2024. **23**(2): p. 141-155.

19. Chandrasekaran, S.N., et al., Image-based profiling for drug discovery: due for a machine-learning upgrade? Nature Reviews Drug Discovery, 2021. **20**(2): p. 145-159.

20. Zhu, H., Big Data and Artificial Intelligence Modeling for Drug Discovery, in Annual Review of Pharmacology and Toxicology, Vol 60, P.A. Insel, Editor. 2020, Annual Reviews: Palo Alto. p. 573-589.

21. Bender, A. and I. Cortés-Ciriano, Artificial intelligence in drug discovery: what is realistic, what are illusions? Part 1: Ways to make an impact, and why we are not there yet. Drug Discovery Today, 2021. **26**(2): p. 511-524.

22. Jiménez-Luna, J., F. Grisoni, and G. Schneider, Drug discovery with explainable artificial intelligence. Nature Machine Intelligence, 2020. **2**(10): p. 573-584.

23. Jiménez-Luna, J., et al., Artificial intelligence in drug discovery: recent advances and future perspectives. Expert Opinion on Drug Discovery, 2021. **16**(9): p. 949-959.

24. Han, R., et al., Revolutionizing Medicinal Chemistry: The Application of Artificial Intelligence (AI) in Early Drug Discovery. Pharmaceuticals, 2023. **16**(9): p. 34.

25. Visan, A.I. and I. Negut, Integrating Artificial Intelligence for Drug Discovery in the Context of Revolutionizing Drug Delivery. Life-Basel, 2024. **14**(2): p. 36.

26. Xie, S.W., et al., A comprehensive analysis of trends in the burden of gout in China and globally from 1990 to 2021. Scientific Reports, 2025. **15**(1): p. 14.

27. Punjwani, S., et al., Burden of gout among different WHO regions, 1990-2019: estimates from the global burden of disease study. Scientific Reports, 2024. **14**(1): p. 12.

28. Li, M.Y., et al., Assessing cross-national inequalities and predictive trends in gout burden: a global perspective (1990-2021). Frontiers in Medicine, 2025. **12**: p. 19.

29. Jeong, Y.J., et al., Global burden of gout in 1990-2019: A systematic analysis of the Global Burden of Disease study 2019. European Journal of Clinical Investigation,